\documentclass[10pt]{article} 
\usepackage{PRIMEarxiv}

\usepackage[utf8]{inputenc} % allow utf-8 input
\usepackage[T1]{fontenc}    % use 8-bit T1 fonts
\usepackage{hyperref}       % hyperlinks
\usepackage{url}            % simple URL typesetting
\usepackage{booktabs}       % professional-quality tables
\usepackage{amsfonts}       % blackboard math symbols
\usepackage{nicefrac}       % compact symbols for 1/2, etc.
\usepackage{microtype}      % microtypography
\usepackage{lipsum}
\usepackage{fancyhdr}       % header
\usepackage{graphicx}       % graphics
\graphicspath{{media/}}     % organize your images and other figures under media/ folder
\usepackage{float}
\usepackage{subcaption}
\usepackage{balance}
\usepackage{multirow}
\usepackage{amssymb}

\let\emptyset\varnothing
\title{
Leveraging Large-scale Multimedia Datasets to Refine Content Moderation Models
}

\author{%
  Ioannis Sarridis \\
  Centre for Research and Technology Hellas\\
  Thessaloniki, Greece\\
  \texttt{gsarridis@iti.gr} \\
  \And
   Christos Koutlis \\
   Centre for Research and Technology Hellas \\
   Thessaloniki, Greece \\
   \texttt{ckoultis@iti.gr} \\
   \And
   Olga Papadopoulou \\
   Centre for Research and Technology Hellas \\
   Thessaloniki, Greece \\
   \texttt{olgapapa@iti.gr} \\
   \And
   Symeon Papadopoulos \\
   Centre for Research and Technology Hellas \\
   Thessaloniki, Greece \\
   \texttt{papadop@iti.gr} \\
}

\begin{document}
\maketitle

\begin{abstract}

The sheer volume of online user-generated content has rendered content moderation technologies essential in order to protect digital platform audiences from content that may cause anxiety, worry, or concern. Despite the efforts towards developing automated solutions to tackle this problem, creating accurate models remains challenging due to the lack of adequate task-specific training data. The fact that manually annotating such data is a highly demanding procedure that could severely affect the annotators' emotional well-being is directly related to the latter limitation. In this paper, we propose the CM-Refinery framework that leverages large-scale multimedia datasets to automatically extend initial training datasets with hard examples that can refine content moderation models, while significantly reducing the involvement of human annotators. We apply our method on two model adaptation strategies designed with respect to the different challenges observed while collecting data, i.e. lack of (i) task-specific negative data or (ii) both positive and negative data. Additionally, we introduce a diversity criterion applied to the data collection process that further enhances the generalization performance of the refined models. The proposed method is evaluated on the Not Safe for Work (NSFW) and disturbing content detection tasks on benchmark datasets achieving 1.32\% and  1.94\% accuracy improvements compared to the state of the art, respectively. Finally, it significantly reduces human involvement, as 92.54\% of data are automatically annotated in case of disturbing content while no human intervention is required for the NSFW task. 
\looseness=-1
\end{abstract}

% keywords can be removed
% \keywords{NSFW \and pornography \and content moderation}

\section{Introduction}
Content moderation is the practice of screening user-generated content (UGC) based on platform-specific rules and guidelines to determine whether a given media item is appropriate for being published on the platform. The reasons why such an item (text, image, video) can be considered inappropriate or harmful  include but are not limited to: violence, offensiveness, extremism, nudity, and hate speech. %, and copyright infringements. 
Concerning text data, hate speech is the most common type of inappropriate content with a growing presence in social media and the internet in general \cite{paz2020hate}. 
Regarding visual data, extensive research has been done on the challenge of Not Safe For Work (NSFW) content that mainly includes nudity and pornography. Although many kinds of NSFW content are acceptable in certain platforms that host exclusively adult content, it is typically restricted from mainstream platforms. %and the internet in general. 
Making the Internet a safe place for people of all ages requires minimizing the possibility of exposure to such data that could adversely affect certain groups of people, such as children.
Additionally, pornography detection can be a valuable tool for law enforcers to detect and restrict many illegal forms of pornography, such as child pornography and revenge porn. Another content moderation task of high importance %but with limited research 
is the detection of disturbing images and videos. Disturbing images and videos are defined as items referring to depictions of humans or animals subjected to violence, harm, and suffering, in a manner that can cause trauma to the viewer \cite{zampoglou2017web}. 

Online platforms employ human annotators to moderate UGC and restrict (typically post hoc) the publishing of content that violates their policies. However, the limited capacity of the human annotation  as well as the psychological distress that can be caused to humans viewing inappropriate content %made infeasible the manual inspection and 
has increasingly turned platforms into largely automated solutions. 
Although, effective content moderation has not yet been fully automated, 
AI-based content moderation systems can reduce the need for human moderation and consequently the impact of viewing harmful content. 

In recent years, Convolutional Neural Networks (CNNs) have demonstrated exceptional performance on various visual tasks, which motivated the research community to automate visual content moderation by employing CNNs. Training a CNN on binary classification tasks requires a considerable amount of positive and negative samples, which is often challenging in relation to content moderation tasks. In the NSFW detection task, there is an abundance of positive data that can be found on the Web, but the collection of a diverse enough negative class is a challenge that leads to CNN models with generalization limitations or many false positives. Moreover, the collection of training data for the disturbing content detection task constitutes an even greater challenge, as even the positive samples are hard to curate given the need for manual inspection; this is reflected in the limited number of such datasets in the literature.

Based on the above, we argue that the main limitation of developing automatic systems for content moderation is the difficulty of collecting large-scale and accurate task-specific training data, while automating the annotation process is of utmost importance as not only manual annotation is non-scalable but it can also result in adverse implications in terms the annotators' emotional well-being.
In order to address these issues, we propose CM-Refinery, a method that leverages large-scale multimedia datasets to refine content moderation models. CM-Refinery considerably reduces the annotators' involvement during data labeling. In particular, we consider two model adaptation strategies with respect to the challenges observed for two different content moderation tasks. The first strategy refers to a task having a comprehensive positive class but a minimally diverse negative class %is the first scenario that 
and is reflected by the NSFW content detection task. 
The second strategy refers to tasks having limited positive samples and again a minimally diverse negative class. This is represented by the disturbing content detection task. 
As regards the NSFW task, after having trained a baseline model, we collect hard negative samples from a large-scale multimedia dataset, namely YFCC100m \cite{thomee2016yfcc100m}, that the baseline model misclassifies. These negative data combined with positive data derived from a task-specific data source, namely the NudeNet dataset \cite{nudenet}, are then utilized to refine the baseline model.
For the disturbing content detection task,
a baseline model is trained and the YFCC100m dataset is utilized to collect data that are considered as positive with a high probability by the baseline model. 
In order to support the auto-labeling process, a manual annotation process for a small number of the derived samples is considered. Then, auto-labeling is achieved by leveraging the cosine similarity between the manually annotated and the remaining unlabeled samples. In addition, we propose a diversity criterion for selecting the most appropriate negative set of samples to further enhance the model's generalization capability. The proposed approach is evaluated on two datasets, one per task of interest, and exhibits superior performance compared to the state of the art. 
%, namely the Pornography-2k \cite{perez2017video} and the Disturbing Images Dataset (DID) \cite{zampoglou2017web}. It is shown to outperform the state of the art in terms of accuracy by 1.32\% (i.e., 97.7\%) and 1.94\% (i.e., 95\%), respectively. 
Finally, it is worth noting that no human annotators were involved in the NSFW task, while for the disturbing content task, the annotators' exposure to such content has been significantly reduced. %, as 92.54\% of the new data was auto-annotated. 
The main contributions of this paper are the following:
\begin{itemize}
\item CM-Refinery, a framework for collecting and annotating task-specific content moderation data, while minimizing the human annotators' involvement (reaching an empirically measured reduction of $\times$13.4) and, hence, their exposure to possibly harmful content. %  An annotation process that minimizes the human annotators' involvement and, respectively, their exposure to possibly harmful content. 
\item  Consideration of two model adaptation strategies with respect to the different data collection and annotation challenges, i.e. lack of (i) task-specific negative data or (ii) both positive and negative data, which are represented by NSFW and disturbing content tasks, respectively. % for collecting and annotating data for refining content moderation models. 
\item A comparative analysis that involves the evaluation of the proposed method on two datasets, namely Pornography-2k \cite{perez2017video} and DID \cite{zampoglou2017web}), in which CM-Refinery achieves 1.32\% and  1.94\% accuracy improvements, respectively. %and a result analysis through class activation maps visualizations for explainability purposes. 
\end{itemize}

% In this paper, motivated by the utmost need of accurate pornography detection models, we explored the reasons that the existing models can not perform well on new real-world data in order to take them into consideration and train a model that not only does it perform well on a specific dataset, but it is also accurate on new real-world data. In particular, we argue that the ineffectiveness of the existing models is due to the training datasets. A major limitation of the existing datasets is that they do not offer the necessary data  diversity, especially for the SFW data. For instance, we expect that a model trained on SFW  data that depict just clothed people will perform worse than a model that is trained on SFW samples that depict people wrestling or swimming.  Taking that into account, we compiled a dataset that consists of samples from several different existing datasets that allows for training an effective model for detecting pornography content in visual data. 

\section{Related work}
\textbf{Content Moderation.} %It is a common practice for 
Social media platforms employ content moderation systems with human annotators/moderators to control whether the UGC violates their policies \cite{gorwa2020algorithmic}. However, the vast amount of UGC led these platforms to turn to Artificial Intelligence (AI) models for moderating large volumes of content and reducing the involvement of human moderators. AI-based content moderation models would need to demonstrate higher performance to replace human moderators entirely. Thus, most of these platforms utilize AI models to make an initial screening of content. Human moderators then review content that has been flagged in order to make the final decision \cite{son2022reliable}. Considering the above, enhancing the performance of AI content moderation models can be a contributory factor to reducing human involvement, which is increasing \cite{arsht2018human} as UGC scales up.

\textbf{NSFW detection.} Early efforts to tackle the task of pornography detection were mainly based on human skin features. In \cite{ap2005algorithm}, the proposed method utilizes appropriate ranges in the RGB, Normalized RGB and HSV colour spaces, to indicate the skin regions in an image and then performs the classification based on the size of the skin regions and their relative distances. A similar method is proposed in \cite{ruiz2005characterizing} that tries to indicate the skin regions and then extract 28 skin-related features for each one of the indicated regions (orientation, size, solidity, position, etc.) that are utilized to train a NSFW classifier. In addition, the authors of \cite{santos2012nudity} argue that pornographic content tends to be placed in the centre of an image; thus, they propose a method that splits images into zones before extracting the skin features. Departing from these methods, \cite{lopes2009bag} introduces a Bag-Of-Features (BOF) approach based on Hue-SIFT descriptors \cite{van2009evaluating} that achieves similar performance without including any shape or geometric modeling. Inspired by this idea, \cite{lopes2009nude} applies a BOF approach for detecting nudity in videos. The discussed traditional approaches share the same shortcomings. First, false positive rates are high, as they are mostly based on the detection of exposed skin. Many types of SFW contexts involve high rates of exposed skin, such as images depicting people sunbathing or fighting in a wrestling match. Second, the existence of pornographic content with a low rate of exposed skin makes those methods unreliable for positive sample detection.

The exceptional performance of the CNNs on visual tasks led researchers to employ them in the context of pornography detection. The first CNN approach, presented in \cite{moustafa2015applying}, fine-tuned the AlexNet \cite{krizhevsky2012imagenet} and  GoogLeNet \cite{szegedy2015going} models on pornography data. Furthermore, \cite{perez2017video} and \cite{geremias2022motion} propose approaches for detecting pornographic content in videos by leveraging both static and motion information. Finally, an effort to tackle the child sexual abuse detection problem is presented in \cite{gangwar2021attm}. In the latter work, the proposed framework consists of two modules: one responsible for pornography detection, and a second that estimates the ages of the depicted people. Although the above CNN-based approaches exhibit high performance on specific datasets, they can not generalize effectively, and thus they face challenges when deployed ``in the wild''.

Regarding training datasets for this task, the abundance of pornography data on the Internet allows for collecting plenty of positive data, which is a widely adopted data collection approach\cite{gangwar2021attm, fu2021pornnet, pandey2021device, yousaf2022deep}. Although collecting positive data is straightforward, populating the negative class with task-oriented data still constitutes a great challenge. Utilizing existing datasets, such as COCO \cite{lin2014microsoft, pandey2021device}, does not provide an effective solution, as such datasets do not offer the data diversity that would enable a network to generalize well. The pornography-2k \cite{perez2017video} constitutes the most well-designed dataset in the literature. It consists of 140 hours of 1000 pornographic and 1000 safe videos. Specifically, Pornography-2k  is an extension of the Pornography-800, or NPDI dataset \cite{avila2013pooling}, and it is characterized by high data diversity. The positive class includes both professional and amateur content, many genres of pornography, several races, and animated content, while the negative class comprises videos of several contexts associated with skin exposure.

\textbf{Disturbing content detection.} %Disturbing content detection constitutes a highly challenging task due to the lack of data. 
Collecting data for disturbing content detection is a great challenge due to the nature of the positive class. % very limited data sources for this task. 
As a result, related datasets and efforts to address this task in the literature are limited. In \cite{IJET22872}, the authors present a CNN based on AlexNet \cite{krizhevsky2012imagenet} as a backbone for detecting and censoring gore images. However, its generalization capability is considerably restricted as the proposed model is trained on only 1000 gore images. Furthermore, \cite{larocque2021gore} presents an effort to address the problem of gore image classification. However, the same limitation applies here, as the constructed dataset consists of 2,097 gore images while including disturbing images derived from movie scenes, which can not effectively substitute the real-world disturbing data. Furthermore, the DID \cite{zampoglou2017web} is another disturbing content dataset that %constitutes a dataset dealing with the challenge of image disturbing content, 
consists of 5401 images, 2043 of which are labeled as disturbing. Although DID can be considered as the largest dataset in the literature, it is still a small-scale dataset. The limitation in terms of scale of the  available datasets prevents the research community from developing highly accurate deep learning-based approaches for addressing the problem of disturbing content detection. In this work, we propose a pipeline to enrich the training data for these tasks and thus enhance the generalization capability of content moderation models. 

\textbf{Human-in-the-loop and interaction aspects.} The nature of the content moderation problem requires highly accurate models. As this is not yet achieved, human-in-the-loop approaches have gained ground. The authors of \cite{link2016human} introduce a framework that supports human moderators' suggestions regarding the relevance and categorization of content for semi-automated content moderation. In addition, an active learning approach for properly selecting the data to be reviewed by human moderators is proposed in \cite{yang2021tar}. In addition, \cite{gorwa2020algorithmic} highlights the need of human-in-the-loop in content moderation as automated moderation models threaten to further increase opacity caused by the insufficient explainability of their decisions. 

However, human annotators' exposure to harmful content constitutes an underappreciated problem of utmost importance. Such daily exposure can cause emotional distress, and severe psychological trauma to the annotators \cite{arsht2018human, steiger2021psychological}. The authors of \cite{das2020fast} propose utilizing image blurring to reduce moderators' exposure to harmful data. However, blurring cannot often convey the information contained in the image that would allow a human moderator to understand what it depicts. Furthermore, \cite{karunakaran2019testing} explores how grayscaling and blurring filters can reduce the emotional impact of content moderation workers. However, they also highlight that the blurred images had usability issues (i.e., image content was unclear, strain was caused on the eyes). These limitations prevent researchers from collecting and annotating such data, which justifies the lack of datasets of adequate size for the disturbing content detection task. In this paper, we propose a method for collecting and annotating such data while considerably reducing human intervention.

\subsection{Problem formulation}
Training a content moderation model for images can be formed as a binary classification problem. Let $f(\cdot)$ denote the baseline model, then $f^{l}(\cdot)$ denotes the model's %$h$ layer output, where $l$ is the model's 
penultimate layer output.
Furthermore, let $D_T$ and $D_L$ denote the task-specific labeled datasets and large multimedia unlabeled datasets, respectively. Note that the latter consists of mainly negative samples. %, i.e. they are considered as ``mostly safe''. 
Even though such datasets can be considered as ``mostly safe'', the fact that they still contain numerous samples of inappropriate content \cite{birhane2021multimodal} is highly considered by CM-Refinery. %Then, the set $\mathcal{I}$ denotes the already labeled data from $D_T$, while 
$\mathcal{M}$, $\mathcal{A}$, and $\mathcal{U}$ denote the manually annotated, automatically annotated, and unlabeled data, respectively, belonging to $D_L$. Also note that, $\mathcal{M} \cup \mathcal{A} \subseteq \mathcal{U}$ and $\mathcal{M} \cap \mathcal{A} = \emptyset$. Finally, the target is to semi-automatically assign labels to $\mathcal{U}$ and expand $D_T$ with both $\mathcal{M}$ and $\mathcal{A}$.

\begin{figure} [t]
    \centering
    \includegraphics[width=0.65\linewidth]{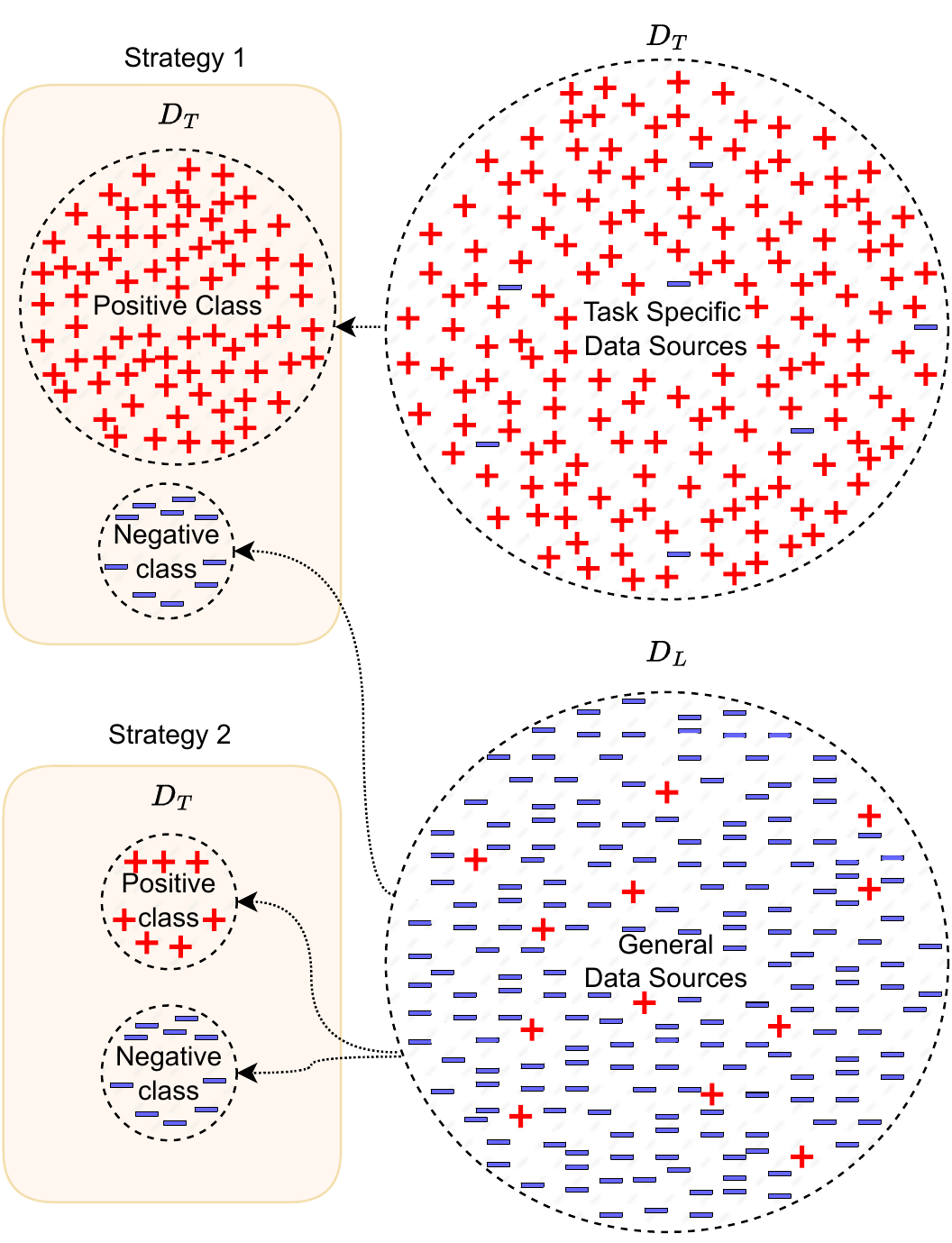}
    \caption{Illustration of the two model adaptation strategies for content moderation. In the first strategy, the positive samples are collected through task-specific data sources, such as websites with pornographic content for the NSFW task, and the negative data from general data sources, such as large multimedia public datasets. In the second strategy, task-specific data sources are unavailable; thus, both positive and negative data are collected from general data sources.}
    \label{fig:scenarios}
\end{figure}
\section{Methodology}

\subsection{Model adaptation strategies}

Here, we define two strategies in terms of the model adaptation process. Note that in this work, the term \textit{data} refers to images unless stated otherwise. The first strategy is related to content moderation tasks where the positive data (e.g. NSFW images) can be easily collected from the web while task-oriented negative samples (e.g. SFW images) should be carefully selected from general data sources, such as large multimedia public datasets (e.g., YFCC100m). For instance, an image that depicts a person in a swimsuit constitutes a task-oriented negative sample for the NSFW task. The second strategy corresponds to tasks that lack data in general (i.e., either positive or negative data). Here, both positive and negative data should be derived from general data sources. For this model adaptation strategy, we experimented with the task of disturbing image detection. Fig. \ref{fig:scenarios} illustrates the model adaptation strategies.

\subsection{Proposed pipeline}
\label{sec:pipeline}
\textbf{Strategy 1}.
The first stage of the proposed pipeline considers training an $f(\cdot)$ model,  for each task. To this end, existing task-specific benchmark datasets are utilized. Given that each task is binary, the binary cross entropy (BCE) loss is employed. %Specifically, the loss function can be defined as follows:
%\begin{equation}
%    \mathcal{L} = -\frac{1}{N}\sum_{i=1}^{N}y_i log(\hat{y}_i) + (1-y_i) log(1-\hat{y}_i),
%\label{eq:bce}
%\end{equation}
%where $N$ denotes the number of samples. 
% The trained baseline models are utilised as feature extractors in the following data collection procedure.
Having selected a large multimedia public dataset $D_L$, the trained baseline models are utilized to classify all the dataset samples and filter those for which the model's predicted score is over a given threshold. Let $\mathbf{X}\in\mathbb{R}^{3\times h\times w}$ denote the input of $f(\cdot)$. Then, this criterion can be defined as follows:
\begin{equation}
    f(\mathbf{X}) > t, ~ t\in [0,1], \forall \mathbf{X} \in D_T
\label{eq:th}
\end{equation}
where $t$ denotes the threshold. The $t$ value should be $>>0.5$ in order to retain those samples that are classified as positive with high probability. Following this procedure, the vast majority of the retained samples are negatives that the baseline model misclassifies. Opting for such hard negative samples to populate the training data, instead of general-purpose data that are trivial to classify, can considerably increase the model's generalization capability.
As regards the positive data for this task, any task-specific data source (e.g., existing datasets, pornographic websites, etc.) can be used for populating the training data. 

\begin{figure} [b]
    \centering
    \includegraphics[width=0.65\linewidth]{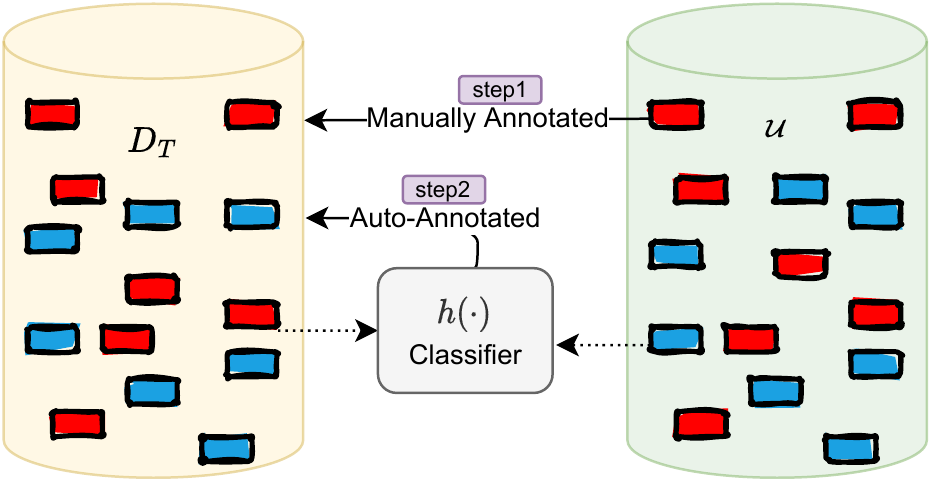}
    \caption{Illustration of the proposed annotation procedure. First, there are two sets of data, the already annotated data from the existing datasets (i.e., $D_T$) and the collected unlabeled data (i.e., $\mathcal{U}$). Second, a few samples of $\mathcal{U}$ are manually annotated. Then, given the similarity criterion, labels are assigned to the remaining unlabeled samples of $\mathcal{U}$ when possible. Finally, both the manually (i.e., $\mathcal{M}$) and the automatically (i.e., $\mathcal{A}$) annotated samples are added to $D_T$.}
    \label{fig:pipeline}
\end{figure}

\textbf{Strategy 2}.
As regards the second model adaptation strategy, which lacks task-specific data sources, the unlabeled data, $\mathcal{U}$, collected from the large multimedia dataset using Eq. \ref{eq:th} are utilized to derive both positive and negative samples. As already mentioned, the majority of these data are negative while only a small portion of them corresponds to positive samples, thus we propose a method for (i) filtering these positive samples and (ii) maximizing the diversity of the selected negative samples. First, a set $\mathcal{M}$ that consists of 1000 samples per class is created by a human annotator, which constitutes the only human input during the proposed procedure. Let $\mathbf{M}_i\in \mathbb{R}^{3 \times h \times w}$ and $\mathbf{U}_j\in \mathbb{R}^{3 \times h \times w}$ denote the $i$-th and the $j$-th sample of $\mathcal{M}$ and $\mathcal{U}$, respectively. Then, the cosine similarity between the manually annotated samples and the remaining unlabeled data is calculated as follows:
\begin{equation}
    s_{i,j} = \frac{ f^l(\mathbf{M}_i) f^l(\mathbf{U}_j)}{||f^l(\mathbf{M}_i)||~||f^l(\mathbf{U}_j)||} 
\end{equation}
%using the baseline model $f^l(\cdot)$ as a feature extractor, the cosine similarity between the features of the manually annotated samples, $\mathcal{M}$, and the features of the remaining unlabeled data, $\mathcal{U}$, is calculated as follows:
% \begin{equation}
%     s_{i,j} = \frac{\mathcal{M}_i \mathcal{U}_j}{||\mathcal{M}_i||~||\mathcal{U}_j||} 
% \end{equation}
Afterwards, a radius near-neighbors (radius-NN) classifier $h(\cdot)$  \cite{bentley1975survey} is utilized to assign labels to $\mathcal{U}$ samples. Let the distance between two samples be calculated as $1-s_{i,j}$ and the radius parameter be denoted as $a$. Here, assigning a low value to $a$ is desirable, as sample pairs of higher similarity are more probable to share the same label. Note that, if more than one labeled samples are in the radius of an unlabeled sample, $j$, the majority vote rule is followed by the radius-NN classifier to assign a label to $j$. Fig. \ref{fig:pipeline} illustrates the annotation procedure.

% and given a threshold $a$, for each $i\in\{0,\cdots,|\mathcal{M}|\}$ and $j\in\{0,\cdots,|\mathcal{U}|\}$, the unlabeled data $\mathbf{U}_j$ are annotated as $y_i$, if the following applies:
% \begin{equation}
%     s_{i,j}>a.
%     \label{eq:th_sim}
% \end{equation}
% Note that assigning a high value to $a$ is desirable, as sample pairs presenting higher similarity, it is more probable to share the same label. Additionally, given a $j$, if Eq. \ref{eq:th_sim} is satisfied for more than one $i$, the majority rule is followed to assign the label to $j$. Figure \ref{fig:pipeline} illustrates the annotation procedure. 

Given that the positive and negative training data should be balanced in terms of their size and the collected positive data in $\mathcal{U}$ are limited, the size of negative data should be selected accordingly.  
Thus, we propose a diversity criterion for maximizing data diversity while selecting the negative data to extend the training data. First, the step of auto-annotating $\mathcal{U}$ data  w.r.t. cosine similarity is repeated once more in order to collect more diverse negative data. 
Having collected a large number of negative data, the target is to opt for the subset that demonstrates the highest diversity. Given that finding the optimal subset in a large-scale set presents extremely high complexity, a random subset generation approach is adopted to find a near-optimal subset. In particular,  $R>>0$ ($R=300,000$ in our experiments) random subsets are created and the standard deviation for each subset is calculated. %Note that in this work, the value of 300,000 is assigned to $R$. 
Then, the subset presenting the higher standard deviation is opted for populating the negative class of data. This way, both the dataset balance and diversity are ensured. After having enriched the training data, the baseline models are retrained to get the final models.

\section{Experimental Setup}
\subsection{Datasets}
The proposed approach has been evaluated on two content moderation tasks, namely NSFW and disturbing content detection. As regards the NSFW detection task, the highly cited Pornography-2k dataset has been used for training the baseline model, while the NudeNetData \cite{nudenet} and the YFCC100m \cite{thomee2016yfcc100m} were used as task-specific and general data sources respectively.  The NudeNetData consists of 483,495 positive samples, while the YFCC100m consists of 99.2 million unlabeled Flickr images. Regarding the disturbing content detection task, the DID \cite{zampoglou2017web} was utilized for training the baseline model and the YFCC100m was considered as the general data source for collecting and labeling new data. Tab. \ref{tab:datasets} presents the described datasets.

\begin{table}[b]
    \centering
    
    \begin{tabular}{|c|c|c|c|c|}
    \hline
        Dataset & Samples & Positive & Negative & Source \\ \hline
        Pornography-2k & 2000 videos & 1000 & 1000 & websites\\  \hline
        NudeNetData & 713,857 images & 483,495 & 230.362 & websites \\  \hline
        \multirow{ 2}{*}{DID} & \multirow{ 2}{*}{5401 images} & \multirow{ 2}{*}{2043} & \multirow{ 2}{*}{3358} & websites \& \\ 
        & & & &  UCID \cite{schaefer2003ucid}\\  \hline
        \multirow{ 2}{*}{YFCC100m} & 99.2M images & \multirow{ 2}{*}{-} & \multirow{ 2}{*}{-} & \multirow{ 2}{*}{Flickr}\\
        & \& 0.8m videos & & & \\  \hline
    \end{tabular} 
    \caption{Basic dataset information for Pornography-2k, NudenetData, DID, and YFCC100m.}
    \label{tab:datasets}
\end{table}

\subsection{Data prepossessing}
Pornography-2k comprises videos, so pre-processing is necessary to extract video frames. In this work, we focus on models for NSFW detection in images, thus, we do not leverage any motion/time information. Considering that in many pornographic videos there are parts that do not depict any NSFW content, centre trimming is applied so that 80\% of their total duration remains. Then, one frame per five seconds is extracted for each video. This procedure resulted in 146,028 SFW and 114,368 NSFW images. Although the centre trimming contributed to properly labeling the frames, we noticed that in some cases, samples labeled as NSFW do not depict any NSFW content. To overcome this, we applied 2-fold (50\%/50\%) cross-validation and manually inspected the false predictions to correct the wrongly labeled samples. The re-annotation procedure resulted in identifying 11,150 samples wrongly labeled as NSFW. Although the human intervention was necessary for the context of this particular dataset pre-processing, this is not part of the proposed method and it does not apply to other datasets.

\subsection{Baseline models}
Selecting a proper CNN architecture for a specific task and dataset is crucial. Here, we opted for the EfficientNet \cite{tan2019efficientnet} models, as they have been shown to be one of the most effective CNN architectures on various visual tasks. EfficientNets allow for scaling their width, depth, and input resolution. This scaling enables us to select a proper variant that fits a given task and dataset. Although the computationally heavy variants (e.g., EfficientNet-b7) outperform the rest on big datasets with many classes, such as ImageNet \cite{deng2009imagenet}, the smaller variants are recommended for either smaller datasets or few classes. Therefore, in this work %For both tasks, 
we selected the EfficientNet-b1 variant. %was selected as network architecture.
% Olga - Taking that into account, since the pornography and image disturbing detection tasks are binary and the training data are <1 million we experimented with the EfficientNet-\{b0-b4\}. Why we do not present these results? 
% Giannis - It refers to the baseline model architecture selection, which is not part of the proposed method we should not focus on this, we just have to mention we opted for b1 variant. In general, you are right, the way I presented in the text makes the reader expect a table with the experiments, I will rephrase it.  

\subsection{Training details and evaluation protocol}
\label{sec:details}
%As regards the models' architecture, several experiments were conducted to opt for the most optimal EfficientNet variant for our purpose. Moreover, for a fair comparison between the variants' performances, several hyperparameter sets were tested. In particular, Adam and SGD optimizers, initial learning rates between 0.0001-0.01, batch sizes 32 and 64, and the learning rate schedule proposed in BiG-Transfer \cite{kolesnikov2020big} were used. As was expected, a small EfficientNet variant, namely EfficientNet-b1, tends to perform better on both content moderation tasks. Thus, EfficientNet-b1 architecture was selected for all the conducted experiments.

As regards the NSFW task, the models are trained using the Stochastic Gradient Decent (SGD) optimizer with 0.9 momentum for 100 epochs with a learning rate of 0.001 and a batch size equal to 32. For the disturbing content detection task, the models are trained for 15 epochs with the Adam optimizer with a learning rate of 0.0001 and for 10 epochs with a learning rate of 0.00001. The batch size is equal to 64. Note that the pretrained weights on the ImageNet dataset were used as initial weights. The threshold hyperparameters $t$ and $a$ defined in Sec. \ref{sec:pipeline} are equal to $0.8$ and $0.85$, respectively. For the threshold $t$, opting for a high value is necessary in order to keep the samples that the baseline model classifies as positive with high confidence while assigning a value close to 1 to $a$ is also desirable as samples with higher similarity sharing the same label with higher probability. The hyperparameter $R$ that denotes the number of random negative subsets (see Sec. \ref{sec:pipeline}) has a value of 300,000. All experiments were conducted using an NVIDIA RTX-3060 GPU. The official test splits were used for the evaluation of the Pornography-2k, while for the evaluation of DID, 80\%/20\% training/test splits were used. Furthermore, to evaluate the generalization capability of the models, 20\% of the data $\mathcal{U}$ derived from YFFC100m was used for testing. The accuracy metric is used to evaluate the classification performance of the models.  Finally, all the class activation maps are derived by the GRAD-CAM approach \cite{selvaraju2017grad}, which is a widely applied method for explaining the decisions of a CNN model.

\section{Results}
\label{sec:res}
Tab. \ref{tab:exp1} presents the results of CM-Refinery for the Pornography-2k dataset compared to five state of the art methods, namely VGG-16 + Bi-RNN \cite{song2020enhanced}, Motion - Optical Flow \cite{perez2017video}, Inter-intra Joint Representation \cite{phan2022joint}, Attention and Metric Learning Based CNN for Pornography (AttM-CNN-Porn) \cite{gangwar2021attm}, and  Frame Sequence Classification (FSC) \cite{gautam2022obscenity}, that leverage both static and motion information. Although we do not take advantage of motion information, our baseline model demonstrates competitive performance compared to the state of the art as demonstrated in Tab. \ref{tab:exp1}. Improving the baseline model by following the proposed pipeline enhances the model's accuracy by 1.32\% (i.e., 97.7\%) on videos, while the accuracy on frames increased from 92.84\% to 95.71\%. It is worth noting that CM-Refinery surpasses the state of the art by 0.55\% in terms of video accuracy while leveraging only static features. Furthermore, for the test set of YFCC100m consisting of samples that the baseline model classifies as positive with confidence (i.e., 0\% accuracy), the proposed model achieves 98.76\% accuracy, which indicates the enhanced generalization capability of the model. 

The results for the DID are presented in Tab. \ref{tab:exp2}. The baseline model achieves 93.06\% accuracy, while the proposed model achieves 95\%. In addition, the accuracy of the proposed model without applying the diversity criterion is 94.44\%, which indicates the need of involving diverse data in the training sets. Finally, the proposed model achieves 79.49\% on the YFCC100m test set, which consists of samples that the baseline model misclassifies.

\begin{table}[b]
    \centering
    % \resizebox{\linewidth}{!}{
    \begin{tabular}{|l|c|c|c|}
    \hline
        \multirow{ 2}{*}{Method}  & \multicolumn{2}{c|}{Pornography-2k} & \multirow{ 2}{*}{YFCC100m}\\ \cline{2-3}
        &   frames & videos & \\ \hline
        VGG-16 + Bi-RNN \cite{song2020enhanced} & - & 95.33\% & - \\ 
        Motion - Optical Flow \cite{perez2017video}	 & - & 96.4\% & - \\ 
        Inter-intra Joint Representation \cite{phan2022joint} & - & 96.88\% & - \\ 
        AttM-CNN-Porn \cite{gangwar2021attm} & - & 97.1\% & - \\ 
        FSC	\cite{gautam2022obscenity} & - & 97.15\% & - \\  \hline
         Baseline (EfficientNet-b1 @ $D_T$) & 92.84\% & 96.38\% & 0\% \\

        % \hspace{1cm} @ Pornography-2k) & & & \\ \hline
         
        CM-Refinery  & \textbf{95.71}\% & \textbf{97.7}\% & 98.76\% \\ \hline
    \end{tabular}
   % }
    \caption{Performance comparison on $D_T$: Pornography-2k.}
    \label{tab:exp1}
\end{table}

\begin{table}[]
    \centering
    \begin{tabular}{|l|c|c|}
    \hline
        Method & DID & YFCC100m \\\hline
        Baseline (EfficientNet-b1 @ $D_T$) & 93.06\% & 0\% \\ \hline
        CM-Refinery (w/o diversity criterion) & 94.44\% & 73.03\% \\ \hline
        CM-Refinery & \textbf{95}\% & 79.49\% \\
        \hline
    \end{tabular}
    \caption{Results of conducted experiments on $D_T$: DID.}
    \label{tab:exp2}
\end{table}
\begin{figure}[]
    \centering
    % \begin{subfigure}[t]{0.3\linewidth}
    %     \centering
    %     \includegraphics[height=1\linewidth]{figures/0.8346_4815429137_929707eb76.jpg}
    %     \caption{Original image.}
    % \end{subfigure}%
    %  ~
    \begin{subfigure}[t]{0.23\linewidth}
        \centering
        \includegraphics[height=1.05\linewidth]{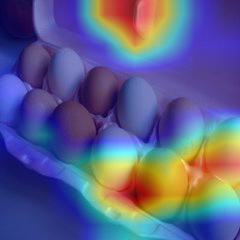}
        \caption{Baseline model: 0.7916}
        \label{fig:eggs_baseline}
    \end{subfigure}
    ~ 
    \begin{subfigure}[t]{0.23\linewidth}
        \centering
        \includegraphics[height=1.05\linewidth]{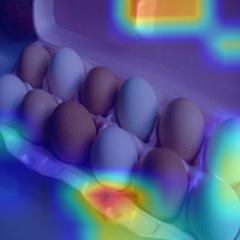}
        \caption{Refined model: 0.0016}
        \label{fig:eggs_tuned}
    \end{subfigure}
    ~
    \begin{subfigure}[t]{0.23\linewidth}
    \centering
    \includegraphics[height=1.05\linewidth]{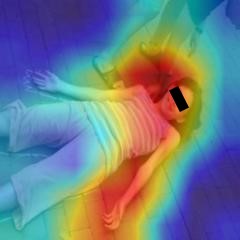}
    \caption{Baseline model: 0.9503}
    \label{fig:ground_baseline}
    \end{subfigure}
     ~ 
    \begin{subfigure}[t]{0.23\linewidth}
        \centering
        \includegraphics[height=1.05\linewidth]{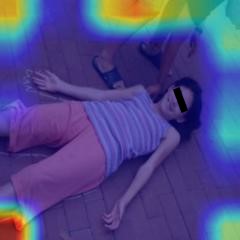}
        \caption{Refined model: 0.005}
        \label{fig:ground_tuned}
    \end{subfigure}
    \caption{Clear samples for humans that can confuse AI models. The baseline model predicts that the image of eggs is NSFW with a score of 0.7916, while the refined model predicts that this image is SFW with a score of 0.0016. Accordingly, the image of a smiling girl that lies on the ground is classified as disturbing with a 0.9503 prediction score by the baseline model, while the refined model classifies it as non-disturbing with high confidence, as the prediction score is equal to 0.005.}
    \label{fig:eggs}
\end{figure}
 \begin{figure}[]
    \centering
    \begin{subfigure}[t]{0.23\linewidth}
        \centering
        \includegraphics[height=1.05\linewidth]{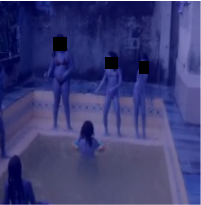}
        \caption{SFW: swimsuits}
        \label{fig:swimsuits}
    
    \end{subfigure}
     ~
    \begin{subfigure}[t]{0.23\linewidth}
        \centering
        \includegraphics[height=1.05\linewidth]{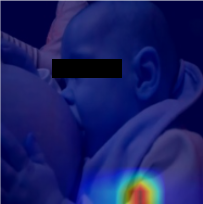}
        \caption{SFW: breastfeeding}
        \label{fig:breastfeeding}
    \end{subfigure}
    ~
    \begin{subfigure}[t]{0.23\linewidth}
    \centering
    \includegraphics[height=1.05\linewidth]{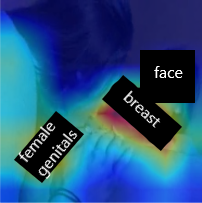}
    \caption{NSFW: female breast and genitals}
    \label{fig:nsfw_female}
    
    \end{subfigure}
     ~
    \begin{subfigure}[t]{0.23\linewidth}
        \centering
        \includegraphics[height=1.05\linewidth]{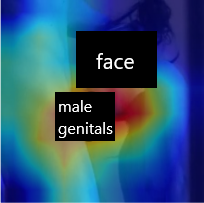}
        \caption{NSFW: male genitals}
        \label{fig:nsfw_male}
    \end{subfigure}
    \caption{Class activation maps for NSFW detection task.}
    \label{fig:nsfw_maps}
\end{figure}
\begin{figure}[t]
    \centering
    \begin{subfigure}[t]{0.23\linewidth}
        \centering
        \includegraphics[height=1.05\linewidth]{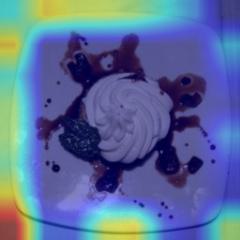}
        \caption{Non-disturbing: dish with red sauce}
        \label{fig:red_sauce}
    
    \end{subfigure}
     ~
    \begin{subfigure}[t]{0.23\linewidth}
        \centering
        \includegraphics[height=1.05\linewidth]{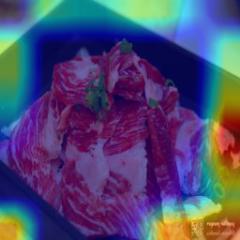}
        \caption{Non-disturbing: raw meat}
        \label{fig:meat}
    \end{subfigure}
    ~
    \begin{subfigure}[t]{0.23\linewidth}
    \centering
    \includegraphics[height=1.05\linewidth]{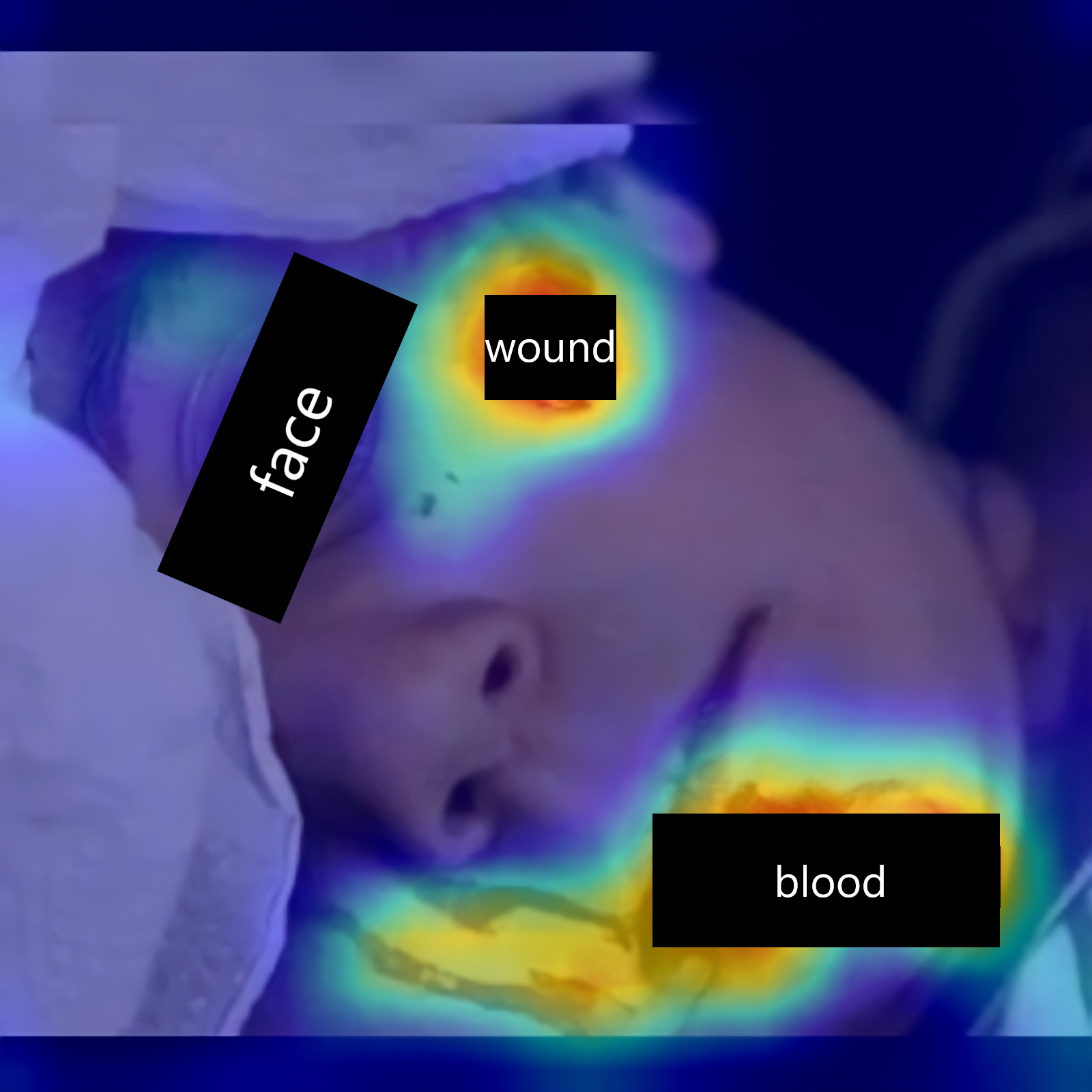}
    \caption{Disturbing: blood}
    \label{fig:blood}
    \end{subfigure}
     ~
    \begin{subfigure}[t]{0.23\linewidth}
        \centering
        \includegraphics[height=1.05\linewidth]{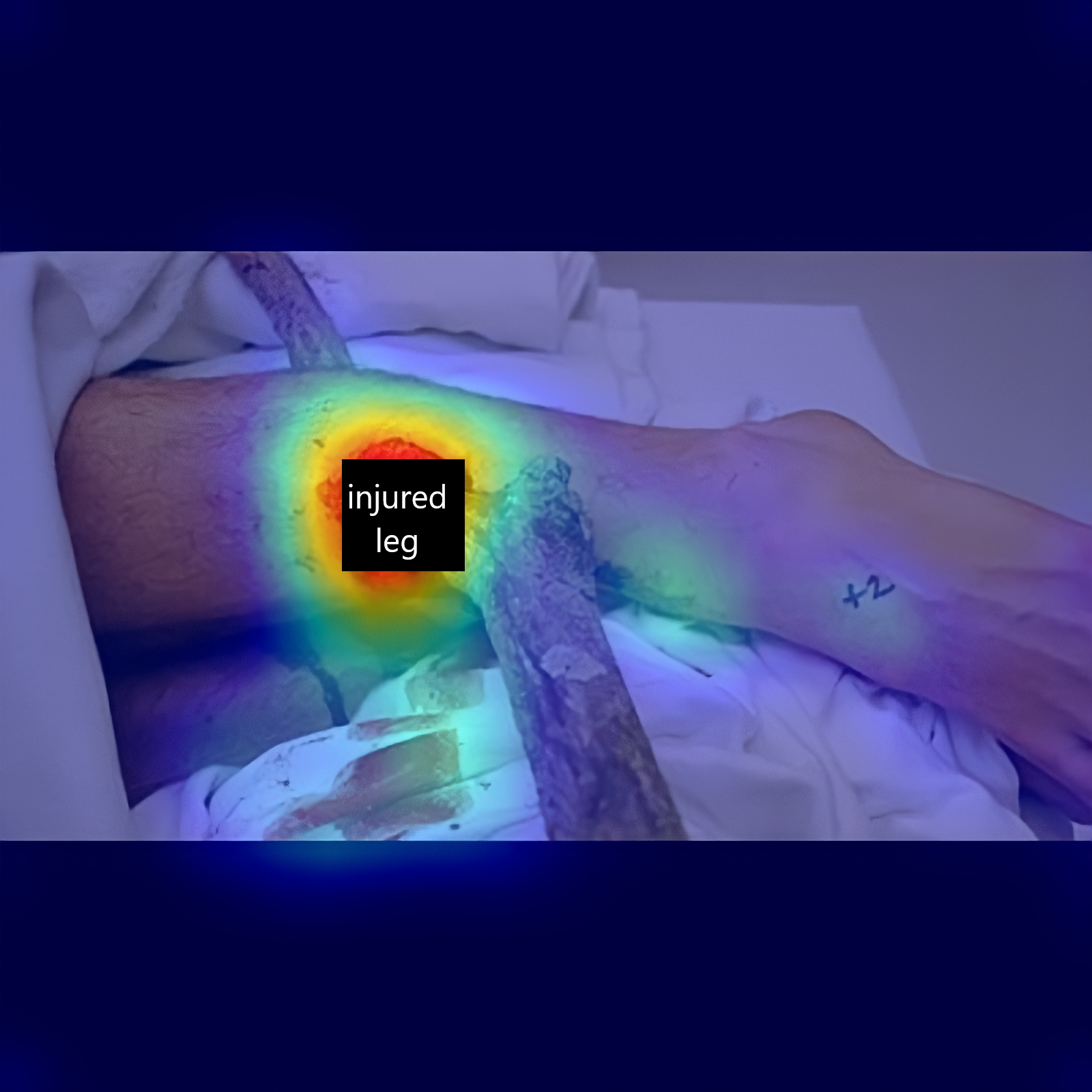}
        \caption{Disturbing: severe wound}
        \label{fig:wound}
    \end{subfigure}
    \caption{Class activation maps for disturbing content detection.}
    \label{fig:dist_maps}
\end{figure}

\subsection{Qualitative analysis}

In a real-world scenario in which users upload images to social media platforms, the biggest proportion of the query images are SFW, and only a small proportion of them are NSFW. Therefore, NSFW detection models should be trained on diverse SFW data to avoid the high rate of false-positives predictions. Although the creators of the Pornography-2k dataset took into consideration the importance of diverse and difficult SFW data, there is one important question that should be discussed: Do human and AI models judge the same samples as challenging? For instance, a human can characterize an image that depicts people wrestling as a challenging sample, which indeed is, but there are other cases that are extremely easy for humans while AI models fail to classify. Fig. \ref{fig:eggs_baseline} and \ref{fig:eggs_tuned} present such an example. Surprisingly, the baseline model fails to classify as SFW a basket of eggs. Although humans ignore its difficulty in being properly classified, it was found to be a very challenging sample for an AI model, as the colour and the edges of the image are quite similar to certain human parts. On the contrary, the refined model can correctly classify this sample as SFW. Accordingly, for the disturbing content detection, there are examples, such as an image of a salad with red sauce or an image that depicts a human lying on the ground, as depicted in Fig. \ref{fig:ground_baseline} and \ref{fig:ground_tuned}. Such limitations are addressed by following the proposed CM-Refinery approach. % for collecting and annotating new data.  

\begin{figure}[t]
        \centering
        \begin{subfigure}[t]{0.22\linewidth}  
            \centering 
            \includegraphics[height=1.05\linewidth]{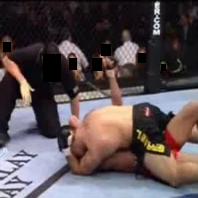}
            \caption[]%
            {{\small DeepAI: 0.9417 (NSFW), Google: 3/5 (NSFW), Ours: 1e-5 (SFW)}}    
            
        \end{subfigure}
        ~
        \begin{subfigure}[t]{0.22\linewidth}   
            \centering 
            \includegraphics[height=1.05\linewidth]{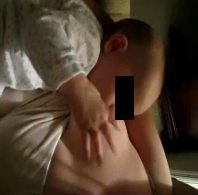}
            \caption[]%
            {{\small DeepAI: 0.9532 (NSFW), Google: 5/5 (NSFW), Ours: 0.1248 (SFW)}}    
            
        \end{subfigure}
        ~
        \begin{subfigure}[t]{0.22\linewidth}   
            \centering 
            \includegraphics[height=1.05\linewidth]{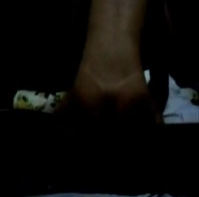}
            \caption[]%
            {{\small DeepAI: 0.0431 (SFW), Google: 4/5 (NSFW), Ours: 0.6845 (NSFW)}}    
            
        \end{subfigure}
        ~
        \begin{subfigure}[t]{0.22\linewidth}   
            \centering 
            \includegraphics[height=1.05\linewidth]{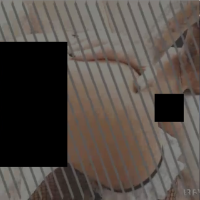}
            \caption[]%
            {{\small  DeepAI: 0.0138 (SFW), Google: 1/5 (SFW), Ours: 0.3453 (SFW)}}    
            \label{fig:comp_mod}
            \end{subfigure}

        \caption[ ]
        {\small CM-Refinery performance on 4 challenging samples (2 NSFW and 2 SFW) compared to Google Vision and DeepAI NSFW Detector services. Google Vision API predictions are on a 1-5 scale (i.e., 1 and 5 correspond to SFW and NSFW with the highest confidence, respectively), while CM-Refinery and DeepAI prediction scores (i.e., [0,1]) denote the model's confidence of an image being NSFW.} 
        \label{fig:comp}
    \end{figure}
Furthermore, the class activation maps for four (two positives and two negatives) samples for the NSFW and disturbing content detection tasks are presented in Fig. \ref{fig:nsfw_maps} and Fig. \ref{fig:dist_maps}, respectively. As can be seen in Fig. \ref{fig:swimsuits} and \ref{fig:breastfeeding}, the refined model correctly classifies the challenging images that depict people wearing swimsuits and breastfeeding as SFW. Furthermore, Fig. \ref{fig:nsfw_female} and \ref{fig:nsfw_male} demonstrate the capability of the refined model to focus on the regions of interest for the NSFW detection task. Accordingly, Fig. \ref{fig:red_sauce} visualizes the class activation map for a sample that depicts a dish with red sauce, and Fig. \ref{fig:meat} demonstrates the activation map for a stack of raw meat. Although both samples are considered very challenging to classify by the disturbing content detection models, the refined model correctly classifies them as non-disturbing. On the other hand, Fig. \ref{fig:blood} and \ref{fig:wound} present the capability of the refined model to focus on the disturbing regions of the images.

Fig. \ref{fig:comp} presents the predictions of the refined NSFW model compared to two commercial NSFW detection services, namely Google Vision AI \footnote{\url{https://cloud.google.com/vision}} and DeepAI NSFW Detection \footnote{\url{https://deepai.org/machine-learning-model/nsfw-detector}}. We opted for 4 (i.e., 2 SFW and 2 NSFW) challenging samples for this purpose.
As negative samples, we opted for images with a high rate of skin depicted (i.e., wrestling, breastfeeding). As can be noticed in Tab. \ref{fig:comp}, both Google Vision AI and DeepAI services consider them wrongly as NSFW, while the CM-Refinery model classifies them correctly. As positive samples, we selected an image with low brightness and an image that has been intentionally modified to confuse the detectors. The DeepAI service fails to detect both the NSFW samples, while the CM-Refinery model and the Google Vision AI fail to detect only the intentionally modified sample (Fig. \ref{fig:comp_mod}).

\section{Conclusions and Future Work}
In this paper, we propose a method for refining content moderation models by leveraging large-scale multimedia datasets. Two model adaptation strategies are considered for two content moderation tasks, namely NSFW and disturbing content detection, as each strategy addresses different challenges while collecting and annotating new data. In addition, the proposed method requires minimal human intervention for data annotation, which is of greatest importance, as such data can have severe effects on the emotional well-being of viewers. The evaluation of the proposed approach shows that not only do the refined models perform better on benchmark datasets, but they also present enhanced generalization capabilities on unknown real-world data. 

As future work, the subjective nature of this task could be explored, as it critically affects the labels of the data and consequently the model's behavior. Strictly defined labeling rules for each content moderation sub-task could assist the development of more reliable and consistent models. Defining fine-grained labels on data could be an additional contributory factor to this end. 

\section*{Acknowledgment}
This research was supported by the EU H2020 project MediaVerse (Grant Agreement 957252).
%Bibliography
\balance
\bibliographystyle{IEEEtran}  
\bibliography{references}

\begin{thebibliography}{10}
\providecommand{\url}[1]{#1}
\csname url@samestyle\endcsname
\providecommand{\newblock}{\relax}
\providecommand{\bibinfo}[2]{#2}
\providecommand{\BIBentrySTDinterwordspacing}{\spaceskip=0pt\relax}
\providecommand{\BIBentryALTinterwordstretchfactor}{4}
\providecommand{\BIBentryALTinterwordspacing}{\spaceskip=\fontdimen2\font plus
\BIBentryALTinterwordstretchfactor\fontdimen3\font minus
  \fontdimen4\font\relax}
\providecommand{\BIBforeignlanguage}[2]{{%
\expandafter\ifx\csname l@#1\endcsname\relax
\typeout{** WARNING: IEEEtran.bst: No hyphenation pattern has been}%
\typeout{** loaded for the language `#1'. Using the pattern for}%
\typeout{** the default language instead.}%
\else
\language=\csname l@#1\endcsname
\fi
#2}}
\providecommand{\BIBdecl}{\relax}
\BIBdecl

\bibitem{paz2020hate}
M.~A. Paz, J.~Montero-D{\'\i}az, and A.~Moreno-Delgado, ``Hate speech: A
  systematized review,'' \emph{Sage Open}, vol.~10, no.~4, p. 2158244020973022,
  2020.

\bibitem{zampoglou2017web}
M.~Zampoglou, S.~Papadopoulos, Y.~Kompatsiaris, and J.~Spangenberg, ``A
  web-based service for disturbing image detection,'' in \emph{International
  Conference on Multimedia Modeling}.\hskip 1em plus 0.5em minus 0.4em\relax
  Springer, 2017, pp. 438--441.

\bibitem{thomee2016yfcc100m}
B.~Thomee, D.~A. Shamma, G.~Friedland, B.~Elizalde, K.~Ni, D.~Poland, D.~Borth,
  and L.-J. Li, ``Yfcc100m: The new data in multimedia research,''
  \emph{Communications of the ACM}, vol.~59, no.~2, pp. 64--73, 2016.

\bibitem{nudenet}
\BIBentryALTinterwordspacing
Archive. (2019) Nudenet dataset. [Online]. Available:
  \url{https://archive.org/details/NudeNet_classifier_dataset_v1}
\BIBentrySTDinterwordspacing

\bibitem{perez2017video}
M.~Perez, S.~Avila, D.~Moreira, D.~Moraes, V.~Testoni, E.~Valle,
  S.~Goldenstein, and A.~Rocha, ``Video pornography detection through deep
  learning techniques and motion information,'' \emph{Neurocomputing}, vol.
  230, pp. 279--293, 2017.

\bibitem{gorwa2020algorithmic}
R.~Gorwa, R.~Binns, and C.~Katzenbach, ``Algorithmic content moderation:
  Technical and political challenges in the automation of platform
  governance,'' \emph{Big Data \& Society}, vol.~7, no.~1, p. 2053951719897945,
  2020.

\bibitem{son2022reliable}
D.~Son, B.~Lew, K.~Choi, Y.~Baek, S.~Choi, B.~Shin, S.~Ha, and B.~Chang,
  ``Reliable decision from multiple subtasks through threshold optimization:
  Content moderation in the wild,'' \emph{arXiv preprint arXiv:2208.07522},
  2022.

\bibitem{arsht2018human}
A.~Arsht and D.~Etcovitch, ``The human cost of online content moderation,''
  \emph{Harvard Journal of Law and Technology}, 2018.

\bibitem{ap2005algorithm}
R.~Ap-Apid, ``An algorithm for nudity detection,'' in \emph{5th Philippine
  Computing Science Congress}, 2005, pp. 201--205.

\bibitem{ruiz2005characterizing}
J.~Ruiz-del Solar, V.~Casta{\~n}eda, R.~Verschae, R.~Baeza-Yates, and F.~Ortiz,
  ``Characterizing objectionable image content (pornography and nude images) of
  specific web segments: Chile as a case study,'' in \emph{Third Latin American
  Web Congress (LA-WEB'2005)}.\hskip 1em plus 0.5em minus 0.4em\relax IEEE,
  2005, pp. 10--pp.

\bibitem{santos2012nudity}
C.~Santos, E.~M. dos Santos, and E.~Souto, ``Nudity detection based on image
  zoning,'' in \emph{2012 11th International Conference on Information Science,
  Signal Processing and their Applications (ISSPA)}.\hskip 1em plus 0.5em minus
  0.4em\relax IEEE, 2012, pp. 1098--1103.

\bibitem{lopes2009bag}
A.~P. Lopes, S.~E. de~Avila, A.~N. Peixoto, R.~S. Oliveira, and A.~d.~A.
  Ara{\'u}jo, ``A bag-of-features approach based on hue-sift descriptor for
  nude detection,'' in \emph{2009 17th European Signal Processing
  Conference}.\hskip 1em plus 0.5em minus 0.4em\relax IEEE, 2009, pp.
  1552--1556.

\bibitem{van2009evaluating}
K.~Van De~Sande, T.~Gevers, and C.~Snoek, ``Evaluating color descriptors for
  object and scene recognition,'' \emph{IEEE transactions on pattern analysis
  and machine intelligence}, vol.~32, no.~9, pp. 1582--1596, 2009.

\bibitem{lopes2009nude}
A.~P.~B. Lopes, S.~E. de~Avila, A.~N. Peixoto, R.~S. Oliveira, M.~d.~M. Coelho,
  and A.~d.~A. Ara{\'u}jo, ``Nude detection in video using
  bag-of-visual-features,'' in \emph{2009 XXII Brazilian Symposium on Computer
  Graphics and Image Processing}.\hskip 1em plus 0.5em minus 0.4em\relax IEEE,
  2009, pp. 224--231.

\bibitem{moustafa2015applying}
M.~Moustafa, ``Applying deep learning to classify pornographic images and
  videos,'' \emph{arXiv preprint arXiv:1511.08899}, 2015.

\bibitem{krizhevsky2012imagenet}
A.~Krizhevsky, I.~Sutskever, and G.~E. Hinton, ``Imagenet classification with
  deep convolutional neural networks,'' \emph{Advances in neural information
  processing systems}, vol.~25, 2012.

\bibitem{szegedy2015going}
C.~Szegedy, W.~Liu, Y.~Jia, P.~Sermanet, S.~Reed, D.~Anguelov, D.~Erhan,
  V.~Vanhoucke, and A.~Rabinovich, ``Going deeper with convolutions,'' in
  \emph{Proceedings of the IEEE conference on computer vision and pattern
  recognition}, 2015, pp. 1--9.

\bibitem{geremias2022motion}
J.~Geremias, E.~K. Viegas, A.~S. Britto~Jr, and A.~O. Santin, ``A motion-based
  approach for real-time detection of pornographic content in videos,'' in
  \emph{Proceedings of the 37th ACM/SIGAPP Symposium on Applied Computing},
  2022, pp. 1066--1073.

\bibitem{gangwar2021attm}
A.~Gangwar, V.~Gonz{\'a}lez-Castro, E.~Alegre, and E.~Fidalgo, ``Attm-cnn:
  Attention and metric learning based cnn for pornography, age and child sexual
  abuse (csa) detection in images,'' \emph{Neurocomputing}, vol. 445, pp.
  81--104, 2021.

\bibitem{fu2021pornnet}
Z.~Fu, J.~Li, G.~Chen, T.~Yu, and T.~Deng, ``Pornnet: a unified deep
  architecture for pornographic video recognition,'' \emph{Applied Sciences},
  vol.~11, no.~7, p. 3066, 2021.

\bibitem{pandey2021device}
A.~Pandey, S.~Moharana, D.~P. Mohanty, A.~Panwar, D.~Agarwal, and S.~P. Thota,
  ``On-device content moderation,'' in \emph{2021 International Joint
  Conference on Neural Networks (IJCNN)}.\hskip 1em plus 0.5em minus
  0.4em\relax IEEE, 2021, pp. 1--7.

\bibitem{yousaf2022deep}
K.~Yousaf and T.~Nawaz, ``A deep learning-based approach for inappropriate
  content detection and classification of youtube videos,'' \emph{IEEE Access},
  vol.~10, pp. 16\,283--16\,298, 2022.

\bibitem{lin2014microsoft}
T.-Y. Lin, M.~Maire, S.~Belongie, J.~Hays, P.~Perona, D.~Ramanan,
  P.~Doll{\'a}r, and C.~L. Zitnick, ``Microsoft coco: Common objects in
  context,'' in \emph{European conference on computer vision}.\hskip 1em plus
  0.5em minus 0.4em\relax Springer, 2014, pp. 740--755.

\bibitem{avila2013pooling}
S.~Avila, N.~Thome, M.~Cord, E.~Valle, and A.~D.~A. Ara{\'u}Jo, ``Pooling in
  image representation: The visual codeword point of view,'' \emph{Computer
  Vision and Image Understanding}, vol. 117, no.~5, pp. 453--465, 2013.

\bibitem{IJET22872}
R.~C. Gustilo, I.~T. Kang, C.~C. So, S.~L. Sy, and A.~S.~I. Crisostomo, ``Gore
  image automatic censorship program,'' \emph{International Journal of
  Engineering \& Technology}, vol.~7, no. 4.16, pp. 138--141, 2018.

\bibitem{larocque2021gore}
W.~Larocque, ``Gore classification and censoring in images,'' Ph.D.
  dissertation, Universit{\'e} d'Ottawa/University of Ottawa, 2021.

\bibitem{link2016human}
D.~Link, B.~Hellingrath, and J.~Ling, ``A human-is-the-loop approach for
  semi-automated content moderation.'' in \emph{ISCRAM}, 2016.

\bibitem{yang2021tar}
E.~Yang, D.~D. Lewis, and O.~Frieder, ``Tar on social media: A framework for
  online content moderation,'' 2021.

\bibitem{steiger2021psychological}
M.~Steiger, T.~J. Bharucha, S.~Venkatagiri, M.~J. Riedl, and M.~Lease, ``The
  psychological well-being of content moderators: the emotional labor of
  commercial moderation and avenues for improving support,'' in
  \emph{Proceedings of the 2021 CHI conference on human factors in computing
  systems}, 2021, pp. 1--14.

\bibitem{das2020fast}
A.~Das, B.~Dang, and M.~Lease, ``Fast, accurate, and healthier: Interactive
  blurring helps moderators reduce exposure to harmful content,'' in
  \emph{Proceedings of the AAAI Conference on Human Computation and
  Crowdsourcing}, vol.~8, 2020, pp. 33--42.

\bibitem{karunakaran2019testing}
S.~Karunakaran and R.~Ramakrishan, ``Testing stylistic interventions to reduce
  emotional impact of content moderation workers,'' in \emph{Proceedings of the
  AAAI Conference on Human Computation and Crowdsourcing}, vol.~7, 2019, pp.
  50--58.

\bibitem{birhane2021multimodal}
A.~Birhane, V.~U. Prabhu, and E.~Kahembwe, ``Multimodal datasets: misogyny,
  pornography, and malignant stereotypes,'' \emph{arXiv preprint
  arXiv:2110.01963}, 2021.

\bibitem{bentley1975survey}
J.~L. Bentley, ``Survey of techniques for fixed radius near neighbor
  searching,'' Stanford Linear Accelerator Center, Calif.(USA), Tech. Rep.,
  1975.

\bibitem{schaefer2003ucid}
G.~Schaefer and M.~Stich, ``Ucid: An uncompressed color image database,'' in
  \emph{Storage and Retrieval Methods and Applications for Multimedia 2004},
  vol. 5307.\hskip 1em plus 0.5em minus 0.4em\relax SPIE, 2003, pp. 472--480.

\bibitem{tan2019efficientnet}
M.~Tan and Q.~Le, ``Efficientnet: Rethinking model scaling for convolutional
  neural networks,'' in \emph{International conference on machine
  learning}.\hskip 1em plus 0.5em minus 0.4em\relax PMLR, 2019, pp. 6105--6114.

\bibitem{deng2009imagenet}
J.~Deng, W.~Dong, R.~Socher, L.-J. Li, K.~Li, and L.~Fei-Fei, ``Imagenet: A
  large-scale hierarchical image database,'' in \emph{2009 IEEE conference on
  computer vision and pattern recognition}.\hskip 1em plus 0.5em minus
  0.4em\relax Ieee, 2009, pp. 248--255.

\bibitem{selvaraju2017grad}
R.~R. Selvaraju, M.~Cogswell, A.~Das, R.~Vedantam, D.~Parikh, and D.~Batra,
  ``Grad-cam: Visual explanations from deep networks via gradient-based
  localization,'' in \emph{Proceedings of the IEEE international conference on
  computer vision}, 2017, pp. 618--626.

\bibitem{song2020enhanced}
K.~Song and Y.-S. Kim, ``An enhanced multimodal stacking scheme for online
  pornographic content detection,'' \emph{Applied Sciences}, vol.~10, no.~8, p.
  2943, 2020.

\bibitem{phan2022joint}
D.-D. Phan, Q.-H. Nguyen, T.-T. Nguyen, H.-L. Tran, and D.-L. Vu, ``Joint
  inter-intra representation learning for pornographic video classification,''
  \emph{Indonesian Journal of Electrical Engineering and Computer Science},
  vol.~25, no.~3, pp. 1481--1488, 2022.

\bibitem{gautam2022obscenity}
N.~Gautam and D.~K. Vishwakarma, ``Obscenity detection in videos through a
  sequential convnet pipeline classifier,'' \emph{IEEE Transactions on
  Cognitive and Developmental Systems}, 2022.

\end{thebibliography}

\end{document}